\documentclass[10pt,twocolumn,letterpaper]{article}

\usepackage[final]{iccv}      

\usepackage{microtype}
\usepackage{multirow}
\usepackage{graphicx}
\usepackage{subcaption}
\usepackage{booktabs} 
\usepackage{bm}

\definecolor{iccvblue}{rgb}{0.21,0.49,0.74}
\usepackage[pagebackref,breaklinks,colorlinks,allcolors=iccvblue]{hyperref}

\usepackage[capitalize]{cleveref}
\crefname{section}{Sec.}{Secs.}
\Crefname{section}{Section}{Sections}
\Crefname{table}{Table}{Tables}
\crefname{table}{Tab.}{Tabs.}
\crefname{equation}{Eqn.}{Eqns.}

\def\paperID{2834} 
\def\confName{ICCV}
\def\confYear{2025}

\title{Can We Achieve Efficient Diffusion without Self-Attention? \\Distilling Self-Attention into Convolutions}

\author{
Ziyi Dong\\
Sun Yat-sen Unviersity\\
{\tt\small dongzy6@mail2.sysu.edu.cn}
\and
Chengxing Zhou\\
Sun Yat-sen Unviersity\\
{\tt\small zhouchx33@mail2.sysu.edu.cn}
\and
Weijian Deng\\
Australian National University\\
{\tt\small dengwj16@gmail.com}
\and
Pengxu Wei\\
Sun Yat-sen Unviersity,\\
Peng Cheng Laboratory\\
{\tt\small weipx3@mail.sysu.edu.cn}
\and
Xiangyang Ji\\
Tsinghua University\\
{\tt\small xyji@tsinghua.edu.cn}
\and
Liang Lin\\
Sun Yat-sen Unviersity,\\
Peng Cheng Laboratory\\
{\tt\small linliang@ieee.org}
}

\begin{document}
\maketitle
\begin{abstract}

Contemporary diffusion models built upon U-Net or Diffusion Transformer (DiT) architectures have revolutionized image generation through transformer-based attention mechanisms. The prevailing paradigm has commonly employed self-attention with quadratic computational complexity to handle global spatial relationships in complex images, thereby synthesizing high-fidelity images with coherent visual semantics. 
{Contrary to conventional wisdom, our systematic layer-wise analysis reveals an interesting discrepancy: self-attention in pre-trained diffusion models predominantly exhibits localized attention patterns, closely resembling convolutional inductive biases. This suggests that global interactions in self-attention may be less critical than commonly assumed.
 Driven by this, we propose \(\Delta\)ConvFusion to replace conventional self-attention modules with Pyramid Convolution Blocks (\(\Delta\)ConvBlocks).
By distilling attention patterns into localized convolutional operations while keeping other components frozen, \(\Delta\)ConvFusion achieves performance comparable to transformer-based counterparts while reducing computational cost by 6929× and surpassing LinFusion by 5.42× in efficiency—all without compromising generative fidelity. } 
\end{abstract}
    
\section{Introduction}

\begin{figure}[t]
    \centering
    \includegraphics[width=0.48\textwidth]{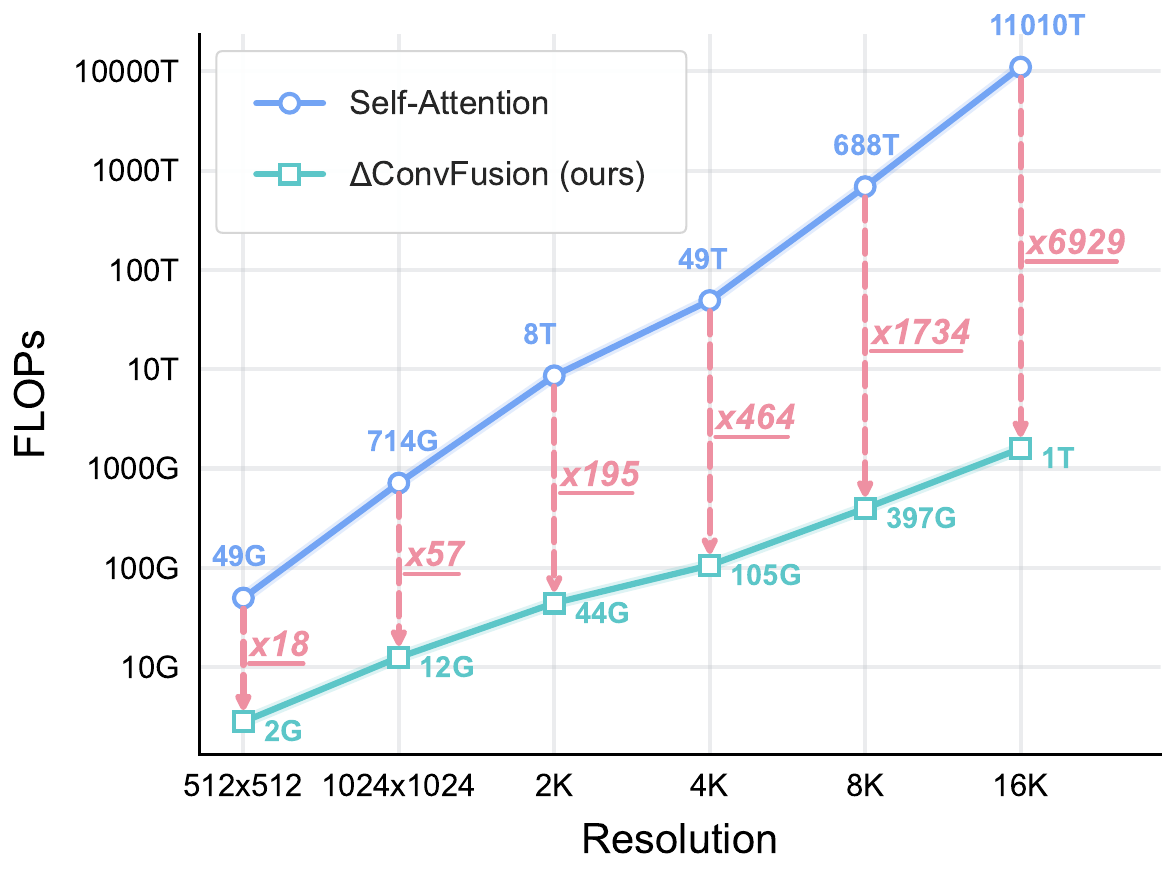}
    \vspace{-20pt}
\caption{Computational cost comparison between self-attention and our \(\Delta\)ConvBlock in SD1.5~\cite{LDM}. Our method significantly reduces FLOPs, achieving up to \(\mathbf{6929\times}\) lower computational cost at 16K resolution while maintaining performance.}\label{fig:flops}
    \vspace{-10pt}
\end{figure}

\begin{figure*}[t]
    \centering
    \includegraphics[width=0.98\textwidth]{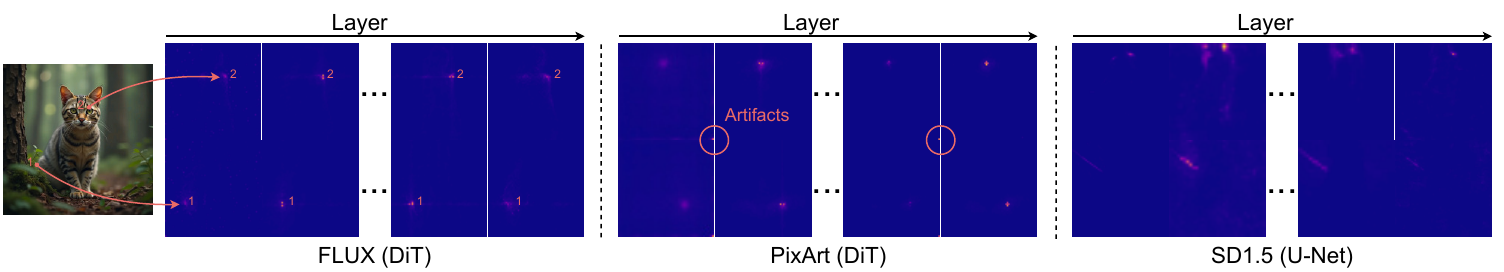}
    \vspace{-6pt}
\caption{Attention visualization across different layers for FLUX, PixArt, and SD1.5. Each column shows attention maps at different layers for a given query location in the input image. The red circles highlight artifacts in PixArt's attention maps (abnormally focus on the last pixel), indicating irregular attention distributions. Across all models, most pixels exhibit highly localized attention, suggesting that self-attention predominantly operates within local neighborhoods rather than modeling global interactions.}
    \label{fig:attn_vis}
\end{figure*}

Image generation has obtained remarkable advancements, with diffusion models~\cite{DDPM,DDIM,ScoreDiff} emerging as a prevalent paradigm that has reshaped the landscape of computer vision research. Contemporary text-conditioned diffusion models~\cite{LDM,ImgGen,SDXL,pixart} excel at generating high-quality images from natural language descriptions, leveraging self-attention mechanisms within U-Net~\cite{UNet,LDM,SDXL,MPDiff} and Diffusion Transformer (DiT)~\cite{DiT,pixart,FLUX} architectures to model spatial and semantic relationships effectively.

Despite significant advancements, a critical challenge has emerged: those diffusion models heavily depend on self-attention that exhibits quadratic computational complexity, especially for high-resolution image generation. As shown in \cref{fig:flops}, self-attention imposes an extremely high computational cost, requiring $11,010$ trillion (T) FLOPs to generate 16K-resolution images. Recent efforts have been devoted to mitigate this issue, such as Mamba-2-based structured state-space models~\cite{LinFusion,Mamba,Mamba2} or windowed attention in selected network layers~\cite{DitFastAttn,swin}. However, these approaches remain constrained by their fundamental assumption of the necessity for global spatial interactions, resulting in suboptimal efficiency improvements and persistent memory inefficiencies.
{
This work considers the following two key questions: \begin{itemize}
    \item {Does self-attention in diffusion models primarily capture global dependencies, or does it behave more locally?}
    \item {Can we effectively replace explicit self-attention computation with a structured convolutional approach while retaining its benefits?}
\end{itemize}
}

In light of the above questions, we systematically investigate the property of self-attention in existing DiT- and U-Net-based diffusion frameworks. 
We first observe that each pixel interacts primarily with its spatial neighbors, forming structured patterns similar to convolutional operations, as shown in \cref{fig:attn_vis}.
Based on this, we further conduct quantitative analysis of self-attention map. The investigation reveals that, despite its theoretical capacity for global interaction, self-attention predominantly exhibits localized patterns. Driven by this, we decompose attention maps into two key components:
(1) a high-frequency, distance-dependent signal, where attention strength decays quadratically with increasing distance from the query pixel, reinforcing its strong locality. 
(2) a low-frequency component, which introduces a spatially invariant bias across the attention map, contributing to broad, spatially smooth attention distribution.

\begin{figure}[t]
    \centering
    \includegraphics[width=0.45\textwidth]{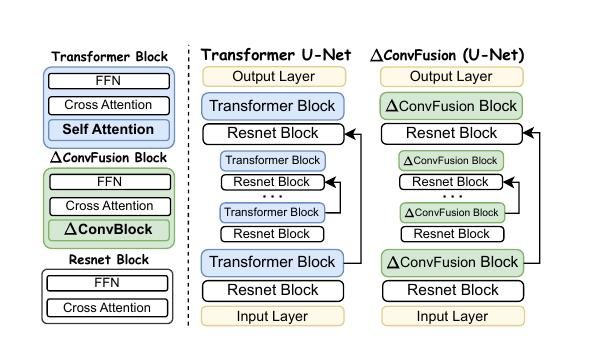}
    \vspace{-2pt}
\caption{Overall diffusion architectures of U-Net and our proposed \(\Delta\)ConvFusion. We replace self-attention modules in Transformer U-Net with \(\Delta\)ConvBlocks, yielding \(\Delta\)ConvFusion (U-Net).}
    \vspace{-6pt}
    \label{fig:overall}
\end{figure}

{Motivated by these insights, we propose {\(\Delta\)ConvFusion}, a novel CNN-based architecture that replaces self-attention with {Pyramid Convolution Blocks} (\(\Delta\)ConvBlocks). Our design leverages multi-scale convolution to efficiently process visual features while replicating the observed attention patterns in diffusion models.  
To align with the two key components of self-attention, \(\Delta\)ConvBlocks incorporate a {pyramid convolutional structure} to capture the high-frequency, distance-dependent signal, while an {average pooling branch} approximates the low-frequency component.
This enables \(\Delta\)ConvBlocks to effectively match the spatial characteristics of self-attention maps.  
We construct \(\Delta\)ConvFusion by replacing self-attention blocks in diffusion models with \(\Delta\)ConvBlocks. 
We show the example on the diffusion architecture of U-Net in \cref{fig:overall}.
To ensure efficient training, we update only \(\Delta\)ConvBlocks while keeping all other components frozen. We further apply knowledge distillation at two levels:  
(1) {feature-level}, minimizing the discrepancy between \(\Delta\)ConvBlocks and the original self-attention modules across all layers; and  
(2) {output-level}, using the \(\epsilon\)-prediction objective~\cite{MinSNR} to align the model’s final outputs.
}

Experimental results demonstrate that $\Delta$ConvFusion can effectively replace self-attention modules in pre-trained diffusion models while achieving comparable or superior image generation quality. 
By distilling self-attention into $\Delta$ConvBlock, our method achieves strong performance across both U-Net and DiT architectures, highlighting its versatility. Compared to existing solutions, $\Delta$ConvFusion significantly reduces computational complexity (FLOPs, as shown in \cref{fig:flops}) and GPU memory consumption while maintaining generation performance. 
$\Delta$ConvBlock achieves 3.4$\times$ speedup (from 1.01$\times$ to 3.4$\times$) over the state-of-the-art LinFusion~\cite{LinFusion} at the  $1024\times1024$ resolution, and 3.37$\times$ speedup (from 96.24$\times$ to 324.1$\times$) at the 16K resolution.
{These findings indicate that explicit global attention mechanisms in diffusion models do not provide benefits proportional to their computational cost, whereas a structured CNN-based approach is sufficient.}

\section{Related Works}

\subsection{Text-to-Image Diffusion Models}
Research in text-to-image synthesis has shifted from GAN-based methods~\cite{sty3,styGANT,ctrlGAN,VQGAN,RiFeGAN} to diffusion models~\cite{DDIM,DDPM,ScoreDiff}. Diffusion models generate high-quality images through an iterative multi-step denoising process and have evolved into large-scale text-to-image frameworks incorporating cross-attention mechanisms~\cite{LDM,GLIDE,ImgGen,SDXL,EDM,pixart,FLUX} and guidance techniques~\cite{guid_cls,guid_cls_free}. By leveraging textual prompts, these models produce visually compelling images, establishing diffusion models as the state-of-the-art approach in text-to-image synthesis.

\subsection{Accelerate Diffusion Models}
Although diffusion models can generate high-quality images, their substantial computational cost remains a major limitation. Recent approaches have explored ways to accelerate diffusion models.  
\cite{DitFastAttn} integrates window attention into selected layers of PixArt~\cite{pixart} (DiT model) to reduce computational overhead. Other works~\cite{DitFastAttn, liu2024faster, chadebec2024flash, selvaraju2024fora, lou2024token, distdiff} aim to optimize computational cost by improving multi-timestep efficiency. \cite{LinFusion} replaces self-attention with structured state-space mechanisms (Mamba-2), significantly reducing the complexity of SD1.5~\cite{LDM} and SDXL~\cite{SDXL} (U-Net model). Additionally, \cite{HDiT} incorporates Neighborhood Attention~\cite{natten} in the shallow layers of U-Net.  
Notably, \cite{distdiff,DitFastAttn, liu2024faster} these methods continue to emphasize the importance of global attention and global modeling capacity in preserving the image generation quality of diffusion models.  

\section{Revisit Self-Attention in Diffusion Models}\label{sec:revisit}

\begin{figure*}[t]
    \centering

    \begin{subfigure}[b]{0.49\textwidth}
        \includegraphics[width=\textwidth]{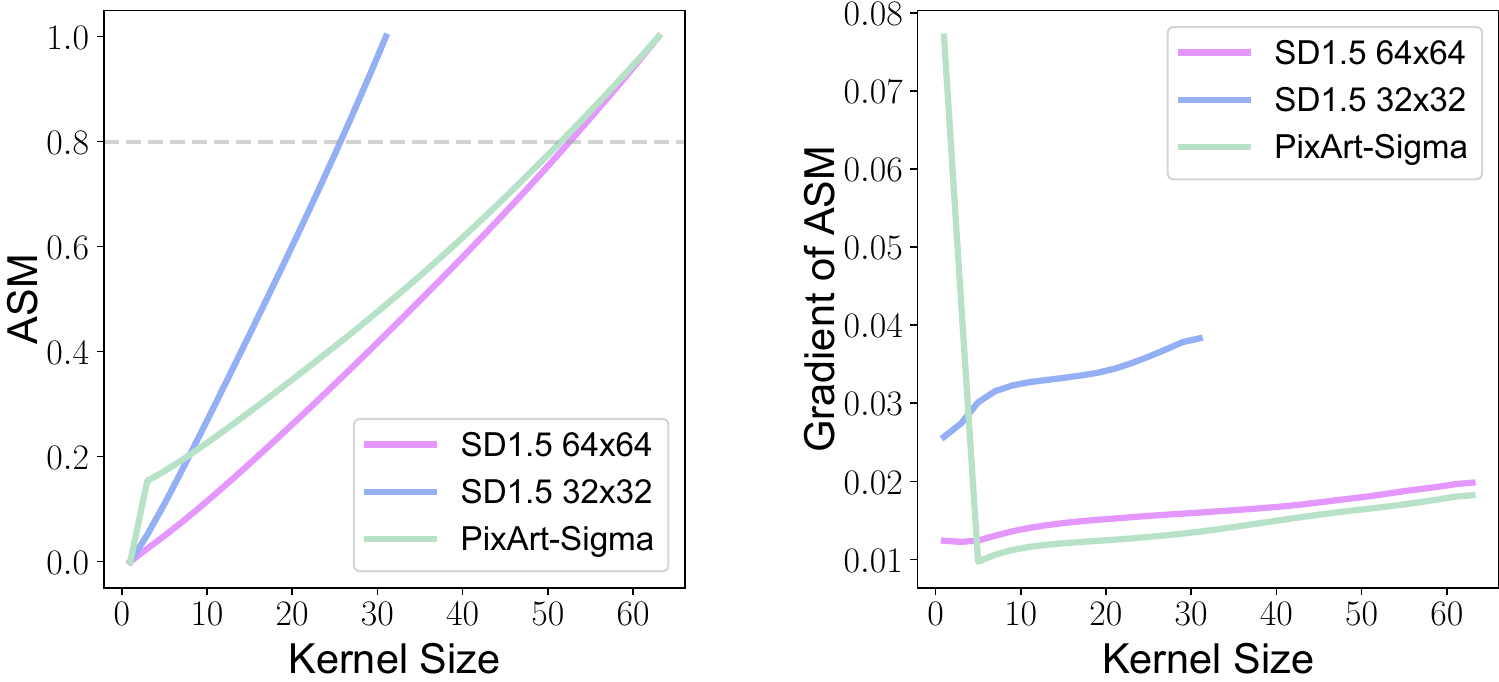}
        \caption{$\mathtt{ASM}$ of whole attention maps.}
        \label{fig:attn:full}
    \end{subfigure}
    \hfill
    \begin{subfigure}[b]{0.49\textwidth}
        \includegraphics[width=\textwidth]{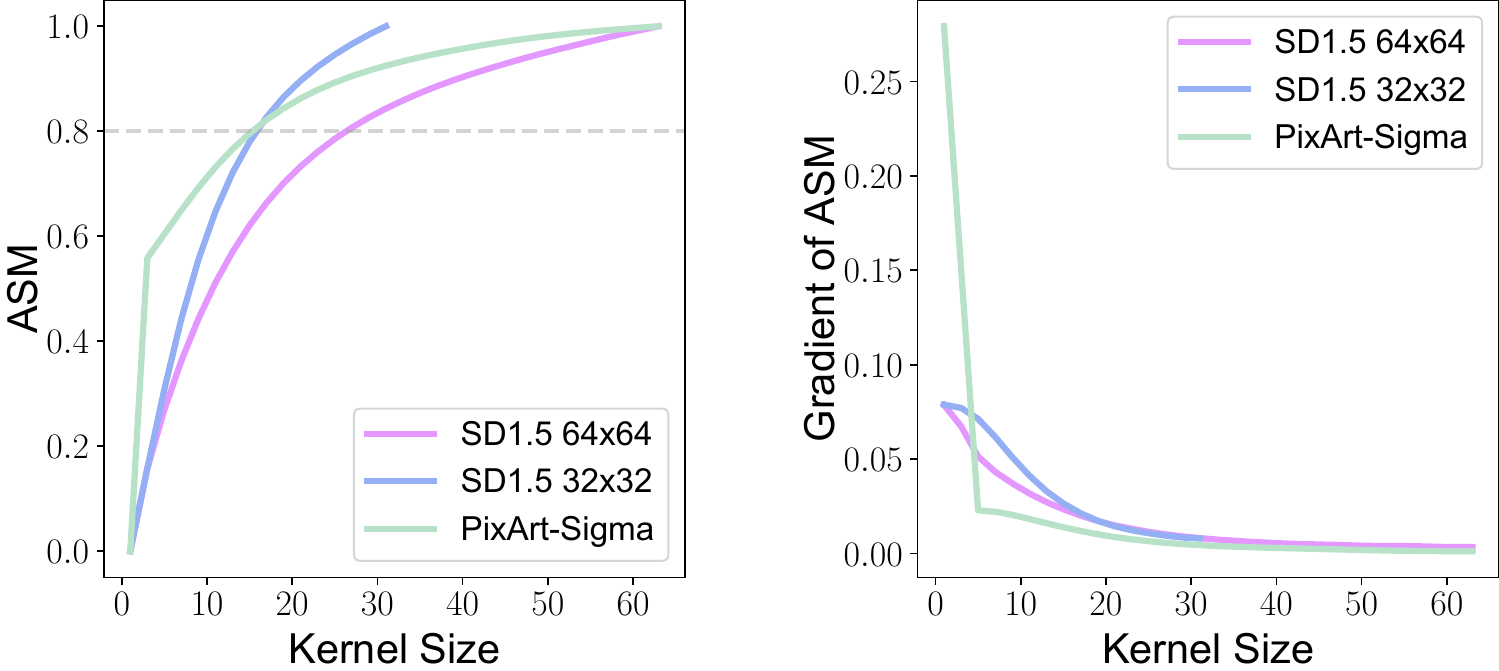}
        \caption{$\mathtt{ASM}$ of high-pass filtered attention maps.}
        \label{fig:attn:fitered}
    \end{subfigure}
    \vspace{-6pt}
    \caption{$\mathtt{ASM}$ and its gradient across kernels with varying sizes. From (a), both PixArt and SD1.5 exhibit a consistent quadratic relationship between kernel size $K$ and $\mathtt{ASM}$. This indicates that attention primarily captures broad low-frequency information, as the accumulated attention follows the typical scaling behavior of spatially smooth signals. From (b), the $\mathtt{ASM}$ of high-pass filtered attention maps decays quadratically with increasing distance from the query pixel, reinforcing its strong locality. {For SD1.5 (U-Net architecture), \(\mathtt{ASM}\) is computed at \(64 \times 64\) and \(32 \times 32\) scales, while PixArt uses a fixed \(64 \times 64\) attention map.}}
    \vspace{-10pt}
\end{figure*}

\subsection{Preliminaries}

Text-to-image diffusion models~\cite{LDM,SDXL,pixart,FLUX} typically consist of two key components: a text encoder~$\mathcal{T}_{\boldsymbol{\phi}}$ and a denoising network~$\epsilon_{\boldsymbol{\delta}}$. Latent Diffusion Models (LDMs)~\cite{LDM} introduce an additional visual compression module, which comprises a visual encoder $\mathbf{z} \gets \mathcal{E}(\mathbf{x})$ and a visual decoder $\mathbf{x} \gets \mathcal{D}(\mathbf{z})$. This module transforms an input image $\mathbf{x}_0 \in \mathbb{R}^{W \times H \times C}$ into a compact latent representation $\mathbf{z}_0 = \mathcal{E}(\mathbf{x}_0) \in \mathbb{R}^{W' \times H' \times C'}$. %
The text prompt $\mathbf{p}$ is encoded by text encoder $\mathcal{T}_{\boldsymbol{\phi}}$ into a semantic embedding~$\mathbf{c} = \mathcal{T}_{\boldsymbol{\phi}}(\mathbf{p})$. The denoising network~$\epsilon_{\boldsymbol{\delta}}$ iteratively refines the visual feature $\mathbf{z}_t$ at the $t$-th diffusion timestep through a time-conditioned denoising process conditioned on $\mathbf{c}$: $\hat{\boldsymbol{\epsilon}} = \epsilon_{\boldsymbol{\delta}}( {\mathbf{z}_t}, \mathbf{c}, t)$, ${\mathbf{z}_t}=\sigma^z_t \mathbf{z}_0 + \sigma^\epsilon_t \boldsymbol{\epsilon}$, 
where $\boldsymbol{\epsilon} \sim \mathcal{N}(\mathbf{0}, \mathbf{I})$ represents Gaussian noise. The noise schedule parameters~$\sigma_t^\epsilon$ and~$\sigma_t^z$ control the magnitude of the added noise and the contribution of original clean data, respectively.

As illustrated in \cref{fig:overall}, diffusion models employing U-Net or DiT architecture conventionally use transformer-based blocks, {where cross-attention facilitates interactions between text embeddings $\bm{c}$ and self-attention enables global spatial interactions across the image.}  

Formally, in the $l$-th attention block in the denoising network, its input with $\mathbf{z}^l_t$ and output $\dot{\mathbf{z}}^l_t$ is defined as: $\dot{\mathbf{z}}^l_t = \mathrm{Softmax} \left( \frac{\boldsymbol{\psi}_q(\mathbf{z}^l_t)^\top \boldsymbol{\psi}_k(\mathbf{z}^l_t)}{\sqrt{C_k}} \right) \boldsymbol{\psi}_v(\mathbf{z}^l_t)$, 
where $\boldsymbol{\psi}_q,\boldsymbol{\psi}_k,\boldsymbol{\psi}_v$ represents the linear layers of $query$, $key$ and $value$ in self-attention, and $\sqrt{C_k}$ is the scaling factor based on the head dimension of self-attention. 
Accordingly, the total FLOPs of self-attention is given by:
$\text{FLOPs}_{\text{attn}} = 4H'W'C'^2 + 4(H'W')^2C' + 4(H'W')^2$,
where $4H'W'C'^2$ is derived from linear layers, $4(H'W')^2C'$ is from matrix multiplication of spatial interactions, and $4(H'W')^2$ is from scaling and softmax. %
As demonstrated by this FLOPs analysis, the computational complexity of self-attention mechanisms exhibits quadratic growth with respect to latent resolution (also scaling proportionally with image resolution), thereby constituting a critical bottleneck in high-resolution image generation tasks.
Recent studies on DiT have demonstrated that the inductive bias inherent in U-Net and convolutional neural networks (CNNs) is not a critical factor for diffusion models to attain state-of-the-art performance~\cite{DiT,pixart}. The superior performance of DiT primarily stems from its explicit global interactions facilitated by self-attention mechanisms. However, this architectural advantage comes at the expense of significantly increased computational overhead. When compared to traditional U-Net based architectures, DiT exhibits substantially higher computational demands, presenting formidable challenges in scaling diffusion models, especially for ultra-high-resolution image synthesis tasks.

\subsection{Our Insight: How Global is Self-Attention?}
\label{sec:self_attn}

{In this section, we investigate the role of global interactions in self-attention and assess their significance in diffusion models. To this end, we systematically analyze the attention patterns in both DiT and U-Net architectural paradigms. 
Additionally, we quantitatively characterize the behavior of self-attention from the frequency domain and further verify its localized patterns through effective receptive field.}

\paragraph{I. Visual Analysis on Self-Attention.}
We analyze the attention maps in self-attention mechanisms to intuitively examine their spatial patterns. Self-attention computes similarity scores between the key representation \(\boldsymbol{\psi}_k(\mathbf{z}^l_t)\) and the query representation \(\boldsymbol{\psi}_q(\mathbf{z}^l_t)\) at the $t$-th timestep, generating an attention map:
$\mathbf{A}^l_t = \mathrm{Softmax} \left( \frac{\boldsymbol{\psi}_q(\mathbf{z}^l_t)^\top \boldsymbol{\psi}_k(\mathbf{z}^l_t)}{\sqrt{C_k}} \right)$,
where \(\mathbf{A}^l_t\) has a shape of \((H'W', H'W')\) and can be reshaped into \((H'W', H', W')\) for visualization.
In \cref{fig:attn_vis}, we present the aggregated self-attention map $\mathbf{A}^l = \frac{1}{T} \sum^T_{t=0} \mathbf{A}^l_t$ across $T$ timesteps in DiT and U-Net architectures, respectively. We observe that almost all the pixels exhibit highly-concentrated attention distributions within their local neighborhoods.

\paragraph{II. Frequency Analysis on Self-Attention.}

Here, we provide a quantitative pattern analysis on self-attention from the frequency domain.
For the \(l\)-th layer, \(\mathbf{A}^l(x_i, y_j)(x_m, y_n)\), one item of $\mathbf{A}^l$, denotes the attention score between the query at \((x_i, y_j)\) and the pixel at \((x_m, y_n)\) in a self-attention map, satisfying:
$\sum_{m=0}^{H'} \sum_{n=0}^{W'} \mathbf{A}^l(x_i, y_j)(x_m, y_n) = 1$.
To measure the self-attention distribution, 
 the {attention score mass} (\(\mathtt{ASM}\)) \((x_i, y_j)\) is defined in a kernel region with kernel size $K$. This kernel is its \(\frac{K}{2}\) \(\ell_\infty\)-neighborhood centered at \((x_i, y_j)\), namely \(d_{\infty} < \frac{K}{2}\):
\begin{equation}
\mathtt{ASM}^l(x_i, y_j) = \sum_{(x_m, y_n) \in d_{\infty} < \frac{K}{2}} \mathbf{A}^l(x_1, y_1)(x_m, y_n),
\end{equation}

To obtain the whole attention score mass for self-attention maps, we then aggregate across spatial positions and all the layers:
$\mathtt{ASM} = \sum_{l} \sum_{i,j=1...H',1...W'} \mathtt{ASM}^l(x_i, y_j)$.
The gradient of \(\mathtt{ASM}\) quantifies the decay rate of attention scores with respect to the increasing spatial distance.

In \cref{fig:attn:full}, both PixArt~\cite{pixart} (DiT architecture) and SD1.5~\cite{LDM} (U-Net architecture) exhibit a consistent quadratic relationship between kernel size \(K\) and \(\mathtt{ASM}\). This indicates that attention primarily captures broad \textit{low-frequency information}, as the accumulated attention score mass follows the typical scaling behavior of spatially smooth signals~\cite{DSP}. 
Moreover, \cref{fig:attn:full} shows that the gradient of \(\mathtt{ASM}\) sharply decreases at small kernel sizes and stabilizes at large kernel sizes, particularly in SD1.5 ($32\times32$), suggesting that most of attentions are concentrated within local neighborhoods. As $K$ increases, the gradients steadily approaches  zero, indicating that expanding the receptive field beyond a certain point does not significantly alter \(\mathtt{ASM}\). 
The analyses above confirms that self-attention, despite its theoretical global capacity, operates mainly in a localized manner in those diffusion models with U-Net and Dit architectures.

To further investigate the role of \textit{high-frequency components}, which encapsulate critical structural information, we perform spectral analysis using a high-pass frequency filtering technique to quantify their characteristics.  
Specifically, for a query at\( (x_i,y_j) \), \( \mathbf{A}^l(x_i,y_j) \) is its self-attention score map with other locations, with the size of \( (H', W') \). It can be regarded as a two-dimensional spatial signal. 
To examine its spectral properties, we apply Discrete Fourier Transform (DFT) and then utilize a high-pass Butterworth filter~\cite{Butterworth1930} to suppress low-frequency components, yielding the high-frequency attention map \( \Lambda^l \). 
We then recompute \(\mathtt{ASM}\) on the high-pass filtered attention map \( \Lambda^l \) as $\mathtt{ASM}^l_{\Lambda}$ to quantify the contribution of high-frequency components.

As shown in \cref{fig:attn:fitered}, for high-pass filtered attention maps, their \(\mathtt{ASM}\) and gradients confirm that most non-trivial attention is concentrated within small local regions, with the gradient attenuating quadratically as the distance increases. In particular, in PixArt, more than 80\% of the high-frequency attention signals are captured within a \(10\!\times\!10\) region, suggesting that self-attention in trained diffusion models does not contribute significantly to long-range interactions.

\begin{figure}[t]
    \centering   \includegraphics[width=0.48\textwidth]{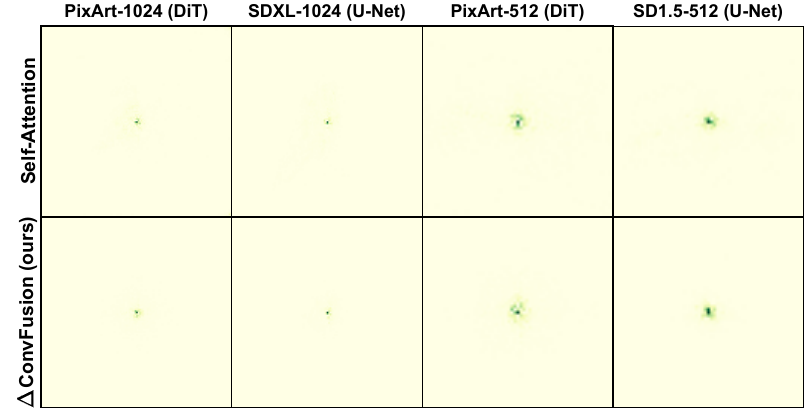}
    \vspace{-18pt}
\caption{Effective Receptive Field (ERF) of self-attention across layers in PixArt-Sigma and SD1.5. PixArt shows larger ERFs in certain layers due to artifacts (Fig.~2) but remains below \(15 \times 15\) in most cases, while SD1.5 stays under \(20 \times 20\). These insights guide \(\Delta\)ConvBlock kernel size selection.}
    \vspace{-12pt}
    \label{fig:erf}
\end{figure}

\paragraph{III. Effective Receptive Field Analysis on Self-Attention.}
{Effective Receptive Field~\cite{ERF} serves as a measure of the effective interaction area, facilitating the design of an appropriate convolutional kernel size as an alternative to self-attention. } 
We define the effective receptive field $\hat{k}^l$ of the $l$-th self-attention as the smallest kernel size over $\Lambda^l$ that encompasses at least 80\% of $\mathtt{ASM}$:
\begin{equation}
    \hat{k}^l = \min\left\{ k \in \{0, \ldots, K\} \,\bigg|\, \mathtt{ASM}^l_{\Lambda} \geq 0.8 \right\},
\end{equation}
As illustrated in \cref{fig:erf}, {the DiT model (PixArt)} exhibits effective receptive fields smaller than 15$\times$15 in most of layers and the U-Net model (SD1.5) exhibits effective receptive fields smaller than 20$\times$20 in most of layers. Notably, the observed larger receptive fields in some transformer blocks (\emph{e.g.}, block-1, block-10 in \cref{fig:erf} left) stem from artifacts in attention maps (\emph{e.g.}, artifacts in \cref{fig:attn_vis}). 
{Self-attention in those transformer blocks are abnormally focus on the last pixel.}
In spite of this phenomenon, their effective attention remains concentrated around query pixels, exhibiting localized spatial patterns.
This further shows that current diffusion models exhibit strong locality, suggesting that their performance does not stem from the global modeling capabilities of self-attention.

\paragraph{Summary of Our Observations.} {The above  analyses provide a support for our hypothesis: while self-attention theoretically enables global interactions, our findings indicate that diffusion models primarily rely on localized patterns during training.  
Furthermore, our results suggest that self-attention can be characterized by two key components:  
(1) A high-frequency, distance-dependent signal, where attention strength decays quadratically with increasing distance from the query pixel, reinforcing its strong locality.  
(2) A low-frequency component, which introduces a spatially invariant bias across the attention map, contributing to broad, spatially smooth attention distribution.}

\subsection{Ablating Global Attention}

\begin{figure}[t]
    \centering
    \includegraphics[width=0.48\textwidth]{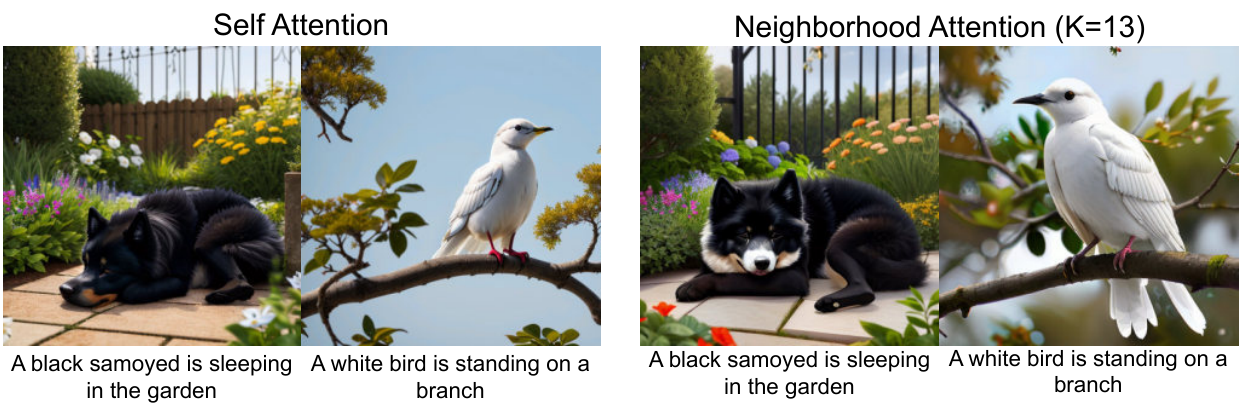}
    \vspace{-18pt}
\caption{Replacing all the self-attention layers in Stable Diffusion with Neighborhood Attention (\(K = 13\)) preserves image quality and semantic coherence, suggesting that global attention is not essential for effective generation.}
    \vspace{-12pt}
    \label{fig:natten_ab}
\end{figure}

Building upon the insights from our observations and quantitative analysis, we systematically investigate the impact of removing global self-attention in diffusion models.

Neighborhood Attention (NA)~\cite{natten} is a localized attention mechanism, for a given pixel at coordinate \((x_1, y_1)\), the query is defined as \(\boldsymbol{\psi}_q(\mathbf{z}^l_t)(x_1, y_1)\). The corresponding {keys} and {values} are strictly confined to a {localized} \(K \times K\) window centered at \((x_1, y_1)\).
Analogous to traditional convolutional layers, NA exhibits localized patterns and receptive field characteristics, making it a viable alternative to global self-attention mechanisms.
As illustrated in {Fig.~\ref{fig:natten_ab}}, replacing {all the self-attention layers} in {Stable Diffusion (SD)} with {NA} (\(K = 13\)) maintains the model's ability to generate semantically coherent and visually high-quality images. 
This further indicates that global interactions may not be necessary at every block in a diffusion model, as self-attention in diffusion models predominantly relies on localized patterns during training. 
However, NA still has high computational overhead and is memory inefficient. Thus, this also motivates us to devise a effective solution for diffusion models in the image generation task.

\section{Our Method: $\Delta$ConvFusion}

\begin{figure*}[!t]
    \centering
    \includegraphics[width=0.98\textwidth]{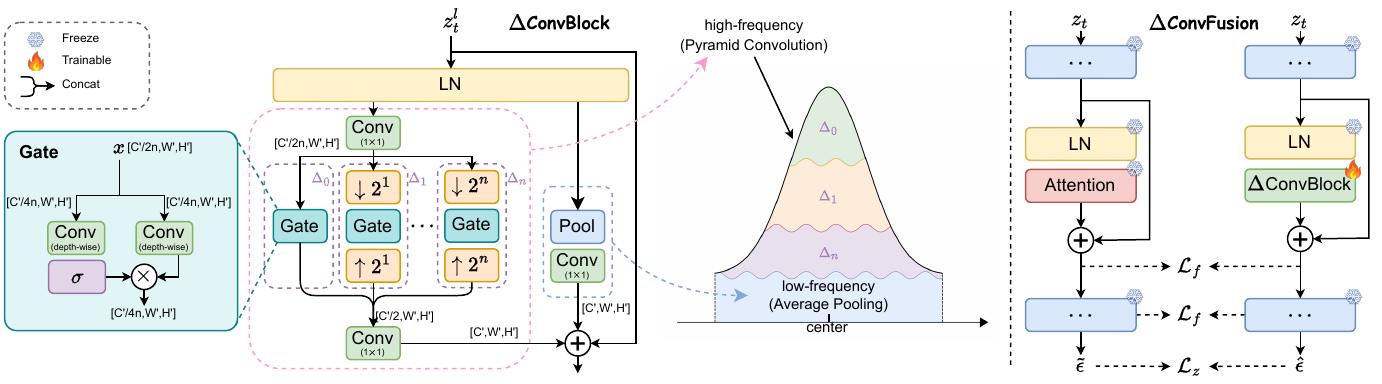}
    \caption{Overview of \(\Delta\)ConvFusion.   
\textbf{Left:} Structure of \(\Delta\)ConvBlock, consisting of two key components:  
    (1) {Pyramid convolution}, which captures high-frequency, distance-dependent features through multi-scale convolutions.  
    (2) {Average pooling}, which models the low-frequency, spatially invariant bias.   
\textbf{Right:} Knowledge distillation framework for \(\Delta\)ConvFusion, where \(\Delta\)ConvBlocks replace self-attention modules while feature-level (\(\mathcal{L}_f\)) and output-level (\(\mathcal{L}_z\)) losses ensure alignment with the original attention-based model.}
    \label{fig:model}
\end{figure*}

The findings in \cref{sec:self_attn} reveal that self-attention in diffusion models operates primarily in a localized manner, reinforcing the potential for more efficient architectures that preserve essential spatial interactions while significantly reducing computational complexity. 
Building on this insight, we propose the \(\Delta\)ConvBlock, a structured convolutional design that effectively captures the core properties of self-attention in diffusion models. Specifically, \(\Delta\)ConvBlock is designed to align with the two key components of self-attention:  
(1) A high-frequency, distance-dependent signal, where attention strength decays quadratically with distance, exhibiting strong locality.  
(2) A low-frequency component/signal, which introduces a spatially invariant bias across the attention map.

\subsection{Pyramid Convolution for Localized Interaction}
To capture the first type of signal, \(\Delta\)ConvBlock incorporates a pyramid convolutional structure with multiple pyramid stages, as illustrated in \cref{fig:model} (left). 

Specifically, given an input latent feature map \(\mathbf{z}^l_t \in \mathbb{R}^{H' \times W' \times C'}\), we first apply {Layer Normalization (LN)} to obtain the normalized feature representation:
$\tilde{\mathbf{z}}^l_t = \text{LN}(\mathbf{z}^l_t).$
Then, an input transformation through a \(1 \times 1\) convolutional layer is applied to facilitate cross-channel interactions while reducing the channel dimension for computational efficiency, defined as
$\mathbf{z}^{l}_{t,\text{in}} = \boldsymbol{\psi}_{\text{in}}(\tilde{\mathbf{z}}^l_t) \in \mathbb{R}^{H' \times W' \times C'/2n},$
where \(\boldsymbol{\psi}_{\text{in}}\) is a  convolutional layer applied to the input, and $n$ is the number of pyramid stages $\{\Delta_0,...,\Delta_n\}$. 
Subsequently, the compressed feature \(\mathbf{z}^{l}_{t,\text{in}}\) is processed via multiple pyramid stages, facilitating hierarchical spatial interactions across multiple scales. At the \(i\)-th pyramid stage, its transformation is 
$\mathbf{z}^{l}_{t,\text{out},i}=\Delta_i(\mathbf{z}^{l}_{t,\text{in}})$, where each stage \(\Delta_i\) incorporates upsampling(\(\uparrow\)) and downsampling (\(\downarrow\)) operations with varying scaling factors, achieving diverse receptive fields, thereby enhancing feature representation.

In one {pyramid stage}, a {scaled simple gate} $\rho(\cdot)$ is implemented with depth-wise convolutions, to enable non-linear spatial interactions. Since the existing simple gate~\cite{NAFNet} employs element-wise multiplication, this design inevitably suffers from numerical overflow under FP16 precision, resulting in training instability. To mitigate this, we introduce a {scaled  simple gate, which enhances training stability and numerical robustness:
$\rho(\mathbf{f}) = \frac{\mathbf{f}_{< C'/2} \cdot \mathbf{f}_{\geq C'/2}}{\sqrt{C'}},$
where $C'$ denotes the number of channels of input feature $\mathbf{f}$.

In this architecture, pixels closer to the center pass through more {pyramid stages}, thereby accumulating higher weights, as illustrated in \cref{fig:model}. This characteristic aligns with the properties of high-frequency signals. 
By leveraging {localized operators} with {linear computational complexity} for multi-scale spatial interactions, our $\Delta$ConvBlock simultaneously enhances local detail preservation and global spatial perception while maintaining a {low computational overhead}:
\begin{equation}
\Delta^l_{\boldsymbol{\theta}}(\mathbf{z}^{l}_{t,\text{in}}) = \sum_{i=1}^{n} \Delta_i(\mathbf{z}^{l}_{t,\text{in}}) = \sum_{i=1}^{n} \left( \uparrow2^i\left(\rho\left(\downarrow2^i(\mathbf{z}^{l}_{t,\text{in}})\right)\right) \right),
\end{equation}
where \(\Delta_i\) is the \(i\)-th pyramid stage, \(\downarrow2^i\) represents the {average pooling layer} with a downsampling scale of \(2^i\) and \(\uparrow2^i\) denotes the {bilinear interpolation operation} with an upsampling scale of \(2^i\).
Finally, a \(1 \times 1\) convolutional layer as the output transformation is also applied.%

\subsection{Average Pooling as Attention Bias}
The second component of \(\Delta\)ConvBlock is an average pooling branch that captures the low-frequency signal. As shown in \cref{fig:model} (blue dashed box), it consists of a \(1 \times 1\) average pooling layer followed by a \(1 \times 1\) convolutional layer \(\boldsymbol{\psi}_p\):
$\mathbf{f}^{\text{avg}}_{\text{out}} = \boldsymbol{\psi}_p \left( \frac{\sum^{W}_{x=0} \sum^{H}_{y=0} \tilde{\mathbf{z}}^l_t(x,y)}{HW} \right).$
Given the linearity of \(1 \times 1\) convolutions and average pooling, this operation satisfies the equivalence:
\begin{equation}
\frac{1}{HW} \boldsymbol{\psi}_p \left( \sum^{W}_{x=0} \sum^{H}_{y=0} \tilde{\mathbf{z}}^l_t(x,y) \right) = \frac{1}{HW} \sum^{W}_{x=0} \sum^{H}_{y=0} \boldsymbol{\psi}_p(\tilde{\mathbf{z}}^l_t(x,y)).
\end{equation}
{The \(1 \times 1\) convolution before and after the average pooling is equivalent.}
This architectural design effectively supplements the low-frequency information captured by self-attention while maintaining {low computational complexity}.

\subsection{Efficient Training via Distillation}

To ensure that \(\Delta\)ConvBlocks can effectively replace self-attention modules in diffusion models while preserving observed locality, we update only \(\Delta\)ConvBlocks and keep all the other parameters frozen. We further apply knowledge distillation to transfer self-attention patterns into \(\Delta\)ConvBlocks, ensuring alignment at both feature and output levels (shown in \cref{fig:model} (right)).
For feature-level alignment, we minimize the discrepancy $\mathcal{L}_f$ between the outputs of the \(\Delta\)ConvBlock and the original self-attention module across all the layers:
\vspace{-5pt}
\begin{equation}
\mathcal{L}_f = \sum_{l=1}^N \left\| \Delta^l_{\boldsymbol{\theta}}(\mathbf{z}^l_t) - \mathbf{z}^l_{t,\text{out}} \right\|^2,
\end{equation}
where \(\Delta^l_{\boldsymbol{\theta}}\) represents the \(l\)-th \(\Delta\)ConvBlock with trainable parameters \(\boldsymbol{\theta}\), and \(\mathbf{z}^l_{t,\text{out}}\) is the output of the \(l\)-th self-attention block.
For output-level alignment, we employ the \(\epsilon\)-prediction objective. To further accelerate convergence, the loss for the output predicts $\mathcal{L}_z$ is improved by {Min-SNR loss weighting}~\cite{MinSNR}:
\vspace{-5pt}
\begin{equation}
\mathcal{L}_z = \min(\gamma \cdot ({\sigma_t^z}/{\sigma_t^\epsilon})^2, 1) \cdot ( \|\tilde{\boldsymbol{\epsilon}} - \hat{\boldsymbol{\epsilon}}\|^2 + \|\boldsymbol{\epsilon} - \hat{\boldsymbol{\epsilon}}\|^2 ),
\end{equation}
where \(\tilde{\boldsymbol{\epsilon}}\) is the output of the self-attention-based model, \(\hat{\boldsymbol{\epsilon}}\) is the output of the \(\Delta\)ConvFusion model, and $\gamma$ is the Min-SNR weighting~\cite{MinSNR}.
Finally, the overall loss function is $\mathcal{L} = \mathcal{L}_z + \beta \mathcal{L}_f$, 
where $\beta$ controls the impact of both losses.

\section{Experiments}

\begin{table*}[!t]
\centering
\caption{Comparison of average computational cost for inference.}
\vspace{-8pt}
\small
\renewcommand\arraystretch{1}
\setlength\tabcolsep{10pt}
\begin{tabular}{cc|rrrrrr}
\hline
\multicolumn{2}{c|}{\multirow{2}{*}{Method}}              & \multicolumn{6}{c}{Mean FLOPs (G) $\downarrow$}                                                                                                                \\
\multicolumn{2}{c|}{}                                         & 512$\times$512 & 1024$\times$1024 & 2K                    & 4K                    & 8K                     & 16K                    \\ \hline
\multirow{4}{*}{PixArt} & Self-Attention~\cite{vaswani2017attention} & 31.42                  & 241.63                  & 2,821.94               & 11,741.69              & 157,546.49              & 2,487,706.88             \\
\multicolumn{1}{c}{}                        & DiTFastAttn~\cite{DitFastAttn}    & 30.41                  & 207.62                  & 2,125.26               & 8,717.07               & 115,012.85              & 1,807,168.71             \\
\multicolumn{1}{c}{}                        & LinFusion~\cite{LinFusion}      & 55.07                  & 220.29                  & 774.46                & 1,858.72               & 6,970.18                & 27,880.73               \\
\rowcolor[HTML]{d0f4de} \multicolumn{1}{c}{}                       & $\Delta$ConvFusion (ours)       & 8.43    & 34.86    & 122.54 & 294.10 & 1,102.85 & 4,411.40 \\
\hline
\multirow{3}{*}{SD1.5}  & Self-Attention~\cite{vaswani2017attention} & 49.67                  & 714.08                  & 8,587.98               & 49,149.21              & 688,821.39              & 11,010,934.11            \\
\multicolumn{1}{l}{}                        & LinFusion~\cite{LinFusion}      & 17.03                  & 68.11                   & 239.43                & 574.64                & 2,154.89                & 8,619.54                \\
\rowcolor[HTML]{d0f4de}\multicolumn{1}{l}{}                         & $\Delta$ConvFusion (ours)       & 2.82    & 12.56    & 44.14  & 105.94 & 397.27  & 1,589.10 \\ \hline
\end{tabular}
\label{tab:infer_cost}
\end{table*}

\subsection{Experimental setup}

\noindent\textbf{Dataset.}
{For attention map analysis in~\cref{sec:self_attn}, we use prompts from the Text2Image-Multi-Prompt dataset~\cite{pszemraj2023text2image}. Our training strategy consists of two phases: first, we pre-train the \(\Delta\)ConvBlock on 2M synthetic images from Midjourney-v5~\cite{mjv5-data}, followed by fine-tuning on 4K curated real images from LAION~\cite{laion2022}, annotated by InternVL2-8B~\cite{chen2024far}, to enhance realism.  
We observe that the widely used MS-COCO benchmark contains a significant number of degraded images, limiting its reliability for high-quality image generation evaluation. To ensure a more rigorous assessment, we evaluate on 10,000 high-aesthetic LAION images, also annotated by InternVL2-8B.}

\noindent\textbf{Implementation details.}
In our experiments, $\Delta$ConvBlock has 2 stages, using $K=13$ for SDXL and PixArt and $K=9$ for SD1.5. In Equ.13, $\gamma$ is set to 5~\cite{MinSNR} and use $\beta=0.001$ to match the loss scales for all the models.
We utilize AdamW optimizer with a learning rate of 3e-4 and $\beta_1=0.9, \beta_2=0.99$ for distillation training. The resolution of input images is 512$\times$512 for SD1.5 and 1024$\times$1024 for SDXL and PixArt, in conjunction with Aspect Ratio Bucket, which automatically groups images of different aspect ratios into different batches and seeks to avoid image cropping as possible. Only $\Delta$ConvBlock needs training and we use a batch size of 16 for training on one A100 GPU. 
For image generation, all experiments employ DPM-Solver~\cite{DPM} sampler with 20 steps and classifier-free guidance (CFG) strength 4.5 for SD1.5, SDXL and PixArt.

\noindent\textbf{Metrics.}
The Inception model~\cite{Incep} has a perceptual discrepancy from human evaluations in evaluating image generation~\cite{stein2023exposing}. For reasonable evaluation, we adopt the DINOv2 model~\cite{dinov2} to evaluate synthesized outputs. Specifically, the DINOv2 Score (DS) quantifies the quality of generated images, while the Fréchet DINOv2 Distance (FDD) measures their realism against a reference dataset. Following~\cite{LinFusion}, the CLIP Score is utilized to evaluate the semantic consistency between generated images and their given text prompts.

\subsection{Computational Complexity and Latency}

\begin{table}[]
\centering
\caption{Comparison of average inference latency.}
\vspace{-8pt}
\small
\renewcommand\arraystretch{1}
\setlength\tabcolsep{3pt}
\begin{tabular}{c|rcrr} %
\hline
\multirow{2}{*}{Method}         & \multicolumn{4}{c}{Inference Latency (ms) $\downarrow$} \\
     & 512$\times$512     & 1024$\times$1024     & 2K        & 4K         \\ \hline
Self-Attention~\cite{vaswani2017attention} & 0.78        & 5.91          & 67.49     & 1,077.98    \\
DiTFastAttn~\cite{DitFastAttn}    & 0.77        & 4.99          & 51.27     & 798.24     \\
LinFusion~\cite{LinFusion}      & 1.52        & 5.82          & 22.36     & 88.59      \\ \hline
 $\Delta$ConvFusion K=13  & 1.09        & 3.53          & 13.40     & 54.84      \\
\rowcolor[HTML]{d0f4de} 
$\Delta$ConvFusion* K=13 & 0.71        & 1.71          & 6.69      & 26.32      \\
$\Delta$ConvFusion K=25  & 2.14        & 8.51          & 33.17     & 131.71     \\
\rowcolor[HTML]{d0f4de} 
$\Delta$ConvFusion* K=25 & 1.23        & 3.79          & 14.73     & 59.95      \\ \hline
\end{tabular}
\label{tab:time}
\end{table}

\begin{table}[]
\centering
\caption{Image generation performance of our $\Delta$ConvFusion and related methods across different base models.}
\vspace{-8pt}
\small
\renewcommand\arraystretch{0.83}
\setlength\tabcolsep{6.5pt}
\begin{tabular}{@{}ccccc@{}}
\toprule
Base Model        & Method         & DS $\uparrow$                     & FDD $\downarrow$                   & CLIP $\uparrow$                 \\ \midrule
\multirow{2}{*}{SD1.5~\cite{LDM}}        & Self-Attention~\cite{vaswani2017attention} & 42.74                  & 210.86                & 30.44                \\
\multirow{2}{*}{512$\times$512}      & LinFusion~\cite{LinFusion}      & 44.23                  & {\textbf{181.78}} & 30.47                \\
             & $\Delta$ConvFusion & {\textbf{44.72}}   & 200.15                & {\textbf{30.73}} \\ \midrule
\multirow{2}{*}{SDXL~\cite{SDXL}}         & Self-Attention~\cite{vaswani2017attention} & 42.65                & { 147.59} & \textbf{30.92}                \\
\multirow{2}{*}{1024$\times$1024}    & LinFusion~\cite{LinFusion}      & 43.24                & 148.61                & 30.78                \\
             & $\Delta$ConvFusion & {\textbf{45.14}} & \textbf{143.72}                & {30.87} \\ \midrule
\multirow{2}{*}{PixArt~\cite{pixart}} & Self-Attention~\cite{vaswani2017attention} & 41.31                  & 173.88                & 30.64                \\
\multirow{2}{*}{512$\times$512}      & DiTFastAttn~\cite{DitFastAttn}    & 41.19                  & \textbf{171.25}                & 30.61                \\
             & $\Delta$ConvFusion &  \textbf{42.56}                      &  181.06                     & \textbf{30.84}                     \\ \midrule
\multirow{2}{*}{PixArt~\cite{pixart}} & Self-Attention~\cite{vaswani2017attention} & 42.74                  & \textbf{180.05}                & \textbf{30.60}                \\
\multirow{2}{*}{1024$\times$1024}    & DiTFastAttn~\cite{DitFastAttn}    & 42.16                  & 184.31                & 30.57                \\
             & $\Delta$ConvFusion &   \textbf{42.95}                     &   181.75                    &   30.55                   \\ \bottomrule
\end{tabular}
\label{tab:quality}
\end{table}

\noindent\textbf{Computational Complexity:}
We quantitatively compare the computational complexity (FLOPs) of our method against state-of-the-art efficient approaches across varying resolutions, as summarized in \cref{tab:infer_cost}. LinFusion~\cite{LinFusion} achieves significant computational cost reduction for resolutions over 4K, exhibiting approximately linear complexity. However, at commonly used resolutions of $512\times512$ and $1024 \times 1024$, LinFusion is not efficient and may even increase computational cost. In contrast, our method substantially reduces computational costs across both low and high resolution regimes while maintaining linear scalability. 
Notably, our approach is orthogonally compatible with AST (Attention Sharing across Timesteps) proposed in \cite{DitFastAttn}, enabling synergistic integration for accelerated inference.

\noindent\textbf{Inference Latency:}
\cref{tab:time} presents the inference latency comparison between our method and existing efficient approaches tested on a GPU. At the widely used 1024$\times$1024 resolution, DiTFastAttn~\cite{DitFastAttn} and LinFusion~\cite{LinFusion} exhibit negligible speed improvements despite their reduced computational complexity. This limitation arises because these approaches still rely on global interaction mechanisms, which are not memory-efficient. In contrast, our method achieves 3.4$\times$ speedup, outperforming existing methods even when employing a 25$\times$25 large convolutional kernel. Furthermore, at high resolutions such as 4K (\emph{e.g.}, 3840$\times$2160), our $\Delta$ConvBlock demonstrates superior inference speed, with an improvement from 12.17$\times$ in LinFusion to 40.96$\times$.

\begin{figure}[!t]
    \centering
    \includegraphics[width=0.48\textwidth]{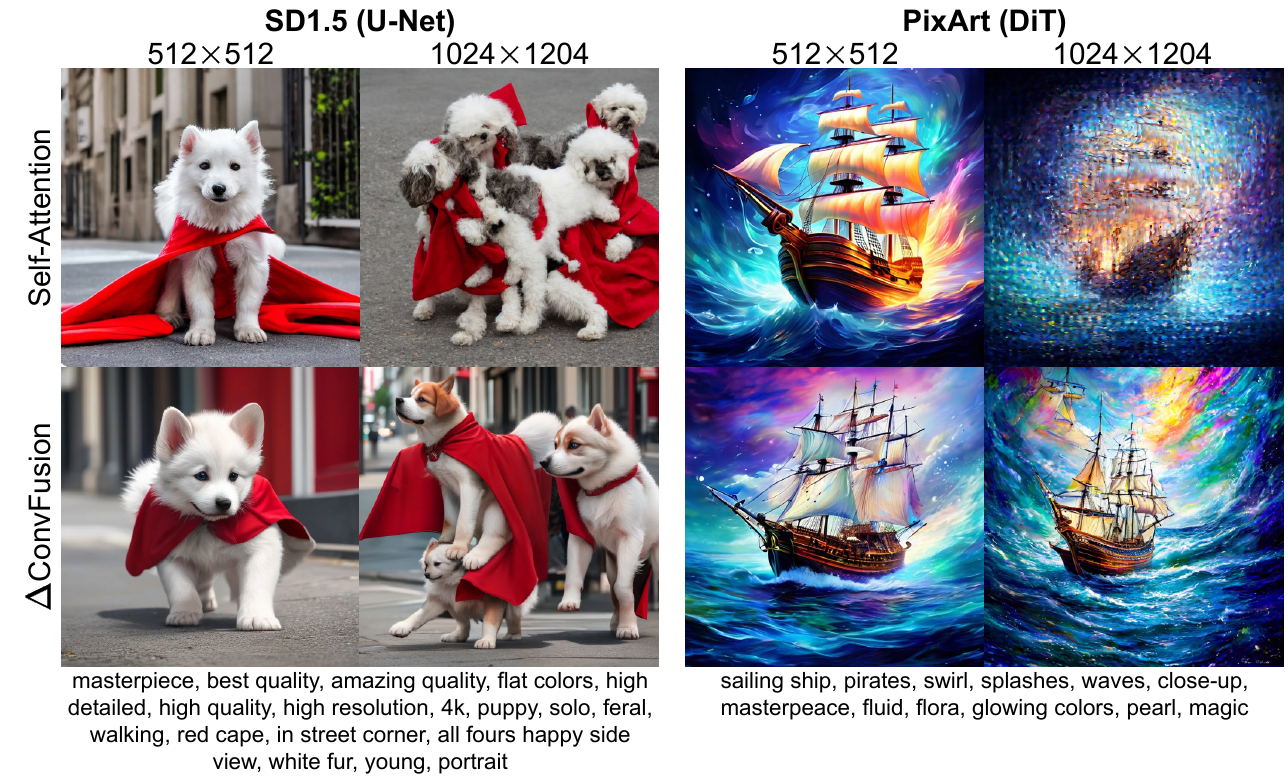}
    \vspace{-18pt}
\caption{Comparison of cross-resolution generation between \(\Delta\)ConvFusion and self-attention-based diffusion models. Both models are trained at \(512 \times 512\) resolution. \(\Delta\)ConvFusion successfully generates coherent \(1024 \times 1024\) images, whereas self-attention-based models produce fragmented and inconsistent features.}  
    \vspace{-5pt}
    \label{fig:corss_res}
\end{figure}

\begin{figure}[!t]
    \centering
    \includegraphics[width=0.48\textwidth]{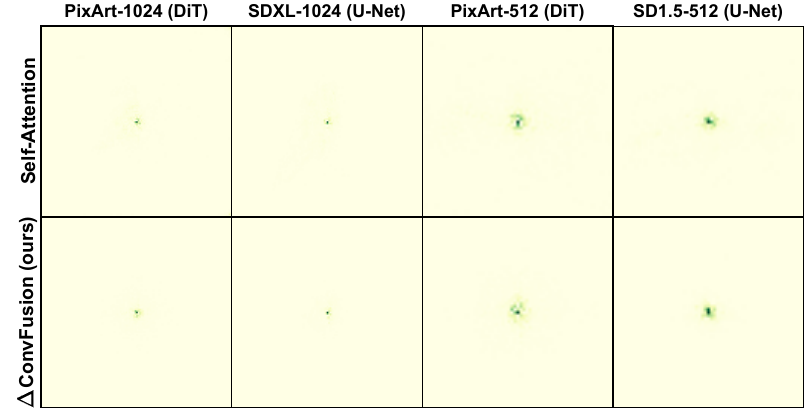}
    \vspace{-18pt}
\caption{Effective Receptive Field (ERF) visualization of \(\Delta\)ConvFusion and self-attention-based diffusion models across SD1.5, SDXL, and PixArt. \(\Delta\)ConvFusion exhibits an ERF pattern closely matching that of self-attention-based models}
    \vspace{-5pt}
    \label{fig:erf_vis}
\end{figure}

\subsection{Qualitative Evaluation}
We replace global self-attention in diffusion models with a linearly structured convolutional module, improving computational efficiency while preserving model performance.  
As shown in \cref{fig:corss_res}, our method produces images that are visually comparable to those generated by self-attention-based diffusion models. In particular, \(\Delta\) ConvFusion improves visual realism and achieves a stronger semantic alignment with textual prompts. Despite relying on local convolutional operations, \(\Delta\)ConvFusion effectively generates semantically coherent images on par with self-attention-based models.

\subsection{Quantitative Evaluation}
We conduct a comprehensive evaluation of
$\Delta$ConvFusion model against existing methods across multiple critical aspects: image fidelity, perceptual quality, diversity, and text-image semantic alignment. 
As shown in \cref{tab:quality}, replacing the self-attention blocks in both DiT and U-Net diffusion models with our $\Delta$ConvBlock achieves superior performance on the DS metric compared to baseline diffusion models and outperforms LinFusion~\cite{LinFusion}. This improvement indicates that our method achieves enhanced image quality and diversity without relying on explicit global interaction mechanisms typically employed in conventional diffusion models. 
Furthermore, as for the FDD metric, the $\Delta$ConvFusion maintain comparable fidelity to their self-attention counterparts.

In terms of text-image semantic alignment for the generation controllability, while $\Delta$ConvFusion solely replaces self-attention modules without tuning cross-attention modules that account for text-image interaction, for the CLIP Score, $\Delta$ConvFusion still achieves a superior alignment between generated images and textual prompts. These empirical results suggest that the hierarchical visual features learned by convolutional 
$\Delta$ConvBlock are more effective for text-to-image generation tasks compared to those derived from self-attention mechanisms. %

\subsection{Model Analysis}
\paragraph{Cross-Resolution Generation:} %
In \cref{fig:corss_res}, \(\Delta\)ConvFusion produces higher-quality images than self-attention-based models when generating resolutions different from the training setting.  
Notably, \(\Delta\)ConvFusion (SD1.5), trained only at \(512 \times 512\), generalizes well to \(1024 \times 1024\), maintaining high visual fidelity.  
In contrast, self-attention-based diffusion models struggle at \(1024 \times 1024\), producing fragmented and chaotic artifacts, such as distorted dogs and ships.  

\paragraph{Effective Receptive Field:}
In \cref{fig:erf_vis}, \(\Delta\)ConvFusion exhibits an ERF pattern closely matching that of self-attention-based models, indicating that \(\Delta\)ConvBlock effectively captures both high- and low-frequency characteristics of attention maps. This strong similarity further validates \(\Delta\)ConvBlock as a computationally efficient yet robust alternative that preserves essential feature extraction capabilities.

\section{Conclusion}
This work systematically analyzes the role of self-attention in diffusion models and studies whether global attention is necessary for effective image generation. Our study reveals that self-attention maps contain two key components: (1) a high frequency, distance-dependent signal, where the attention strength decays quadratically with increasing distance from the query pixel, exhibiting strong locality.  
(2) A low-frequency global component, which manifests as a spatially invariant attention bias across the feature map.  
Based on this, we propose \(\Delta\)ConvBlock, a novel module that decouples these complementary mechanisms through a dual-branch design:  
(1) Pyramid convolution operators capture localized high-frequency patterns using multi-scale receptive fields.  
(2) Adaptive average pooling layers approximate the global low-frequency bias.  
This structured decomposition enables \(\Delta\)ConvBlock to preserve the functional properties of self-attention while reducing its quadratic computational complexity. Extensive experiments across DiT and U-Net-based diffusion models demonstrate that our approach maintains generation quality while significantly improving efficiency.

\newpage
{
    \small
    \bibliographystyle{ieeenat_fullname}
    \bibliography{main}
}

\end{document}